\documentclass{article}

\usepackage{microtype}
\usepackage{graphicx}
\usepackage{booktabs} 
\usepackage{amsmath}

\usepackage{epsfig} 
\usepackage{graphicx} 
\usepackage{dsfont}
\usepackage{amssymb}
\usepackage{subcaption}

\usepackage{hyperref}

\usepackage[accepted]{icml2018}

\icmltitlerunning{Long-term Video Prediction without Supervision}

\begin{document}

\twocolumn[
\icmltitle{Hierarchical Long-term Video Prediction without Supervision}




\begin{icmlauthorlist}
\icmlauthor{Nevan Wichers}{goo}
\icmlauthor{Ruben Villegas}{umich}
\icmlauthor{Dumitru Erhan}{goo}
\icmlauthor{Honglak Lee}{goo} 
\end{icmlauthorlist}

\icmlaffiliation{umich}{Department of Electrical Engineering and Computer Science, University of Michigan, Ann Arbor, MI, USA.}
\icmlaffiliation{goo}{Google Brain, Mountain View, CA, USA.}

\icmlcorrespondingauthor{Nevan Wichers}{wichersn@google.com}

\icmlkeywords{Machine Learning, ICML, Video Prediction, Long term Video Prediction}

\vskip 0.3in
]

\iftrue  
        \newcommand{\todo}[1]{}
        \newcommand{\outline}[1]{}
        \newcommand{\textgray}[1]{}
        \newcommand{\commenttext}[1]{}
        \newcommand{\commentfoot}[1]{}
        \newcommand{\commentselfoot}[2]{}
        \newcommand{\topic}[1]{}
\else 
        \newcommand{\todo}[1]{{\textcolor{red}{[[TODO: {#1}]]}}}
        \newcommand{\outline}[1]{{\textcolor{blue}{[[{#1}]]}}}
        \newcommand{\textgray}[1]{\textcolor{gray}{[[{#1}]]}}
        \newcommand{\commenttext}[1]{\textcolor{red}{[[{#1}]]}}
        \newcommand{\commentfoot}[1]{\footnote{\textcolor{red}{\textit{#1}}}}
        \newcommand{\commentselfoot}[2]{{\textcolor{blue}{#1}}\commenttext{#2}}
        \newcommand{\topic}[1]{\textcolor{gray}{\textbf{(#1.)}}}
\fi


\iftrue 
        \newcommand{\cutsectionup}{\vspace*{-0.06in}}
        \newcommand{\cutsectiondown}{\vspace*{-0.05in}}

        \newcommand{\cutsubsectionup}{\vspace*{-0.06in}}
        \newcommand{\cutsubsectiondown}{\vspace*{-0.05in}}

        \newcommand{\cutsubsubsectionup}{\vspace*{-0.05in}}
        \newcommand{\cutsubsubsectiondown}{\vspace*{-0.05in}}

        \newcommand{\cutparagraphup}{\vspace*{-0.05in}}
        \newcommand{\cutparagraphdown}{\vspace*{-0in}}

        \newcommand{\cutcaptionup}{\vspace*{-0.0in}}
        \newcommand{\cutcaptiondown}{\vspace*{-0.0in}}

        \newcommand{\cuttablecaptionup}{\vspace*{-0.0in}}
        \newcommand{\cuttablecaptiondown}{\vspace*{-0.0in}}

        \newcommand{\cutequationup}{\vspace*{-0.0in}}
        \newcommand{\cutequationdown}{\vspace*{-0.0in}}

        \newcommand{\cuttableup}{}
        \newcommand{\cuttabledown}{}

        \newcommand{\cut}{{\vspace*{-0.02in}}}
        \newcommand{\cutmore}{{\vspace*{-0.06in}}}
        \newcommand{\negcut}{}
\else 
        \newcommand{\cutsectionup}{}
        \newcommand{\cutsectiondown}{}

        \newcommand{\cutsubsectionup}{}
        \newcommand{\cutsubsectiondown}{}

        \newcommand{\cutsubsubsectionup}{}
        \newcommand{\cutsubsubsectiondown}{}

        \newcommand{\cutparagraphup}{\vspace*{-0.05in}}
        \newcommand{\cutparagraphdown}{}

        \newcommand{\cutcaptionup}{}
        \newcommand{\cutcaptiondown}{}

        \newcommand{\cutequationup}{}
        \newcommand{\cutequationdown}{}

        \newcommand{\cuttableup}{}
        \newcommand{\cuttabledown}{}

        \newcommand{\cut}{}
        \newcommand{\cutmore}{}
        \newcommand{\negcut}{}
\fi



\printAffiliationsAndNotice{}  

\begin{abstract}
Much of recent research has been devoted to video prediction and generation, yet most of the previous works have demonstrated only limited success in generating videos on short-term horizons.
The hierarchical video prediction method by \citet{2017arXiv170405831V} is an example of a state-of-the-art method for long-term video prediction, but their method is limited because it requires ground truth annotation of high-level structures (e.g., human joint landmarks) at training time. 
Our network encodes the input frame, predicts a high-level encoding into the future, and then a decoder with access to the first frame produces the predicted image from the predicted encoding. 
The decoder also produces a mask that outlines the predicted foreground object (e.g., person) as a by-product.
Unlike \citet{2017arXiv170405831V}, we develop a novel training method that jointly trains the encoder, the predictor, and the decoder together without high-level supervision; we further improve upon this by using an adversarial loss in the feature space to train the predictor.
Our method can predict about 20 seconds into the future and provides better results compared to \citet{Denton} and \citet{2016arXiv160507157F} on the Human 3.6M dataset.
\end{abstract}
\cutsectionup
\section{Introduction}
\cutsectiondown

Building a model that is able to predict the future states of an environment from raw high-dimensional sensory data (e.g., video) has recently emerged as an important research problem in machine learning and computer vision. 
Models that are able to accurately predict the future can play a vital role in developing intelligent agents that interact with their environment~\cite{jayaraman2015learning,jayaraman2016look,2016arXiv160507157F}.

Popular video prediction approaches focus on recursively observing the generated frames to make predictions farther into the future~\cite{Oh15,Mathieu16,Goroshin15,Srivastava15,Ranzato14,2016arXiv160507157F,VillegasICLR17,Lotter17}.
In order to make reasonable long-term frame predictions in natural videos, these approaches need to automatically identify the dynamics of the main factors of variation changing through time, while also being highly robust to pixel-level noise.
However, it is common for the previously mentioned methods to generate quality predictions for the first few steps, but then the prediction dramatically degrades until all of the video context is lost or the predicted motion becomes static.

A hierarchical method makes predictions in a high-level information hierarchy (e.g., landmarks) and then decodes the predicted future in high-level back into low-level pixel space.
The advantage of predicting the future in high-level space first is that the predictions degrade less quickly compared to predictions made solely in pixel space. 
The method by~\citet{2017arXiv170405831V} is an example of a hierarchical model; however, it requires ground truth human landmark annotations during training time.
In this work, we explore ways to generate videos using a hierarchical model \emph{without} requiring ground truth landmarks or other high-level structure annotations during training.
In a similar fashion to~\citet{2017arXiv170405831V}, the proposed network predicts the pixels of future video frames given the first few frames.
Specifically, our network never observes any of the predicted frames, and the predicted future frames are driven solely by the high-level space predictions.

The contributions of our work are summarized below:
\begin{itemize}
\item An unsupervised approach for discovering high-level features necessary for long-term future prediction.
\item A joint training strategy for generating high-level features from low-level features and low-level features from high-level features simultaneously.
\item Use of adversarial training in feature space for improved high-level feature discovery and generation.
\item Long-term pixel-level video prediction for about 20 seconds into the future for the Human 3.6M dataset.
\end{itemize}

\cutsectionup
\section{Related Work}
\cutsectiondown

\paragraph{Patch-level prediction}
The video prediction problem was initially studied at the patch level containing synthetic motions~\citep{NIPS2008_3567,michalski_grammar_cells,icml2014c2_mittelman14}.
\citet{Srivastava15} and~\citet{Ranzato14} followed up by proposing methods that can handle prediction in natural videos. However, predicting patches encounters the well-known aperture problem that causes blockiness as prediction advances in time.

\cutparagraphup
\paragraph{Frame-level prediction on realistic videos.}
More recently, the video prediction problem has been formulated at the full frame level using convolutional encoder/decoder networks as the main component.
\citet{2016arXiv160507157F} proposed a network that can perform next frame video prediction by explicitly predicting pixel movement. For each pixel in the previous frame, the network outputs a distribution over locations that a pixel is predicted to move. The possible movement a pixel can make are then averaged to obtain the final prediction. The network is trained end-to-end to minimize L2 loss.
\citet{Mathieu16} proposed adversarial training with multiscale convolutional networks to generate sharper pixel-level predictions in comparison to the conventional L2 loss.
\citet{2017arXiv170405831V} proposed a network that decomposes motion and content in video prediction and obtained more accurate results over \citet{Mathieu16}.
\citet{Lotter17} proposed a deep predictive coding network in which each layer learns to predict the lower-level difference between the future frame and current frame.
As an alternative approach to convolutional encoder-decoder networks, \citet{vpn} proposed an autoregressive generation scheme for improved prediction performance. 
In a concurrent work, \citet{DBLP:journals/corr/abs-1710-11252} and \citet{Denton} proposed stochastic video prediction method based on recurrent variational autoencoders. 
Despite these efforts, long-term prediction on high-resolution natural videos beyond approximately $20$ frames has been known to be very challenging.

\cutparagraphup
\paragraph{Long-term prediction.}
\citet{Oh15} proposed an action conditional convolutional encoder-decoder architecture that demonstrated high-quality long-term prediction performance on video games (e.g., Atari games), but it has not been applied to real-world video prediction.
\citet{2017arXiv170405831V} proposed a long-term prediction method using a hierarchical approach, but it requires the ground truth landmarks as supervision. 
Our work proposes several techniques to address this limitation.

\cutsectionup
\section{Background}
\cutsectiondown

The hierarchical video prediction model in ~\citet{2017arXiv170405831V} relieves the blurring problem observed in previous prediction approaches by modeling the video dynamics in high-level feature space.
This approach enables the prediction of many frames into the future.
The hierarchical prediction model is described below.

To predict the image at timestep $t$, the following procedure is used:
First, the high-level features $p_t \in \mathds{R}^l$ \textemdash \ in this case human pose landmarks \textemdash \ are estimated from the first $C$ context frames.
Next, an LSTM is used to predict the future landmark states $\hat{p}_t \in \mathds{R}^l$ given the landmarks estimated from the context frames as follows:
\begin{equation}
\begin{cases} 
    \left[ \hat{p}_{t}, H_{t} \right] = \mathit{LSTM}( p_{t-1}, H_{t-1}) & \text{if} \ t \leq C, \\
    \left[ \hat{p}_{t}, H_{t} \right] = \mathit{LSTM}( \hat{p}_{t-1}, H_{t-1}) & \text{if} \ t > C,
\end{cases} \nonumber
\end{equation} 
where $H_{t} \in \mathds{R}^h$ is the hidden state of the LSTM at timestep $t$.
Note that the predicted $\hat{p}_{t}$ after $C$ timesteps is used to generate the video frames.
Additionally, they remove the auto-regressive connections that feed $\hat{p}_{t-1}$ back into LSTM making the prediction only depend on $H_{t-1}$.
In our formulation, however, the prediction depends on both $\hat{p}_{t-1}$ and $H_{t-1}$, but $\hat{p}_{t-1}$ is not a vector of landmarks.

Once all $\hat{p}_{t}$ are obtained, the visual analogy network (VAN)~\citep{NIPS2015_5845} generates the corresponding image at time $t$.
VAN identifies the change between $g(p_C)$ and $g(\hat{p}_t)$, where $g(.)$ is a fixed function that takes in landmarks and converts them into Gaussian heatmaps.
Next, it applies the identified difference to image $I_C$ to generate image $I_t$.
The VAN does this by mapping images to a space where analogies can be represented by additions and subtractions.
Therefore, the image at timestep $t$ is computed by
\begin{align}
    & \hat{I}_t = \mathit{VAN}(p_{C}, \hat{p}_{t}, I_{C}) = \nonumber \\
    & f_{dec}( \, f_{\mathit{pose}}(g(\hat{p}_{t})) - f_{\mathit{pose}}(g(p_{C})) + f_{img}(I_{C})  \, ). \nonumber
\end{align}
In contrast to \citet{2017arXiv170405831V}, our method does not require landmarks $p_{t}$, and therefore the dependence on the fixed function $g(.)$ is removed.
Our method automatically discovers the features needed as input to the VAN for generating frame at time $t$.
These features locate the object moving through time, and help our network focus on generating the moving object pixels in future frames.
In the following section, we describe our method and training variations for unsupervised future frame prediction.

\cutsectionup
\section{Method}
\label{sec:proposed_method}
\cutsectiondown

\subsection{Network Architecture}
\cutsubsectiondown

Our method uses a network architecture similar to \citet{2017arXiv170405831V}. However, our predictor LSTM and VAN do not require landmark annotations and can be trained jointly.
In our model, the predictor LSTM is defined by
\begin{equation}
\begin{cases} 
    \left[ \hat{e}_{t}, H_{t} \right] = \mathit{LSTM}( e_{t-1}, H_{t-1} ) & \text{if} \ t \leq C \\
    \left[ \hat{e}_{t}, H_{t} \right] = \mathit{LSTM}( \hat{e}_{t-1}, H_{t-1} ) & \text{if} \ t > C, \\
\end{cases}
\end{equation}
where $e_{t-1} \in \mathds{R}^d$ is a general feature vector computed from an input image $I_t$ by an encoder network, and $\hat{e}_t \in \mathds{R}^d$ is the feature vector predicted by the LSTM.
To compute the frame at time $t$, we use a variation of the \textit{deep} version of the image analogy formulation from~\citet{NIPS2015_5845}. 
In contrast to~\citet{2017arXiv170405831V}, we use the first frame in the input video to compute the future frames via image analogy.
Therefore, the frame at time $t$ is computed by
\begin{align}
    &\bar{I}_t, M_t = \mathit{VAN}(e_{1}, \hat{e}_{t}, I_{1}) = \nonumber \\
    &f_{dec}( \, f_{enc}(\hat{e}_{t}) + T(f_{img}(I_{1}), f_{enc}(e_{1}), f_{enc}(\hat{e}_{t}))  \, ),  \\
    &\hat{I}_t \qquad = \bar{I}_t \odot M_t + (1 - M_t) \odot I_1, \label{eq:mask}
\end{align}
where $f_{enc}:\mathds{R}^d \rightarrow \mathds{R}^{s \times s \times m}$ is a convolutional network that maps a feature vector into a feature tensor, $f_{img}: \mathds{R}^{h \times w \times c} \rightarrow \mathds{R}^{s \times s \times m}$ is a convolutional network that maps an input image into a feature tensor, $f_{dec}: \mathds{R}^{s \times s \times m} \rightarrow \mathds{R}^{h \times w \times c}$ is a deconvolutional network that maps a feature tensor into an image, and $T(.,.,.)$ is defined as follows:
\begin{align}
    &T(x,y,z) = f_{analogy}([f_{diff}(x-y),z]),
\end{align}
where $f_{diff}: \mathds{R}^{s \times s \times m} \rightarrow \mathds{R}^{s \times s \times m}$ computes a feature tensor from the difference between $x$ and $y$, $[.,.]$ denotes a concatenation along the depth dimension of the input tensors, and $f_{analogy}: \mathds{R}^{s \times s \times 2m} \rightarrow \mathds{R}^{s \times s \times m}$ computes the analogy feature tensor to be added to $f_{enc}(\hat{e}_{t})$.
Finally, $M_t$ is a gating mechanism that enables our network to identify the moving objects in the video frames.
In Equation~\ref{eq:mask}, our network chooses pixels from the input frame that can simply be copied into the predicted frame, and pixels that need to be generated are chosen from $\bar{I}_t$.
In Section~\ref{sec:experiments}, we show that the selected areas resemble the structure of moving objects in the input and the predicted frames.

\cutsubsectionup
\subsection{Training Objective}
\cutsubsectiondown

These networks can be trained in multiple ways.
In \citet{2017arXiv170405831V}, the predictor LSTM and VAN are trained separately using ground truth landmarks.
In this work, we explore alternative ways of training these networks in the absence of ground truth annotations of high-level structures.

\cutsubsubsectionup
\subsubsection{End-to-End Prediction}
\label{sec:e2e}
\cutsubsubsectiondown

\begin{figure}[t]
\begin{center}
\includegraphics[width=1.0\linewidth]{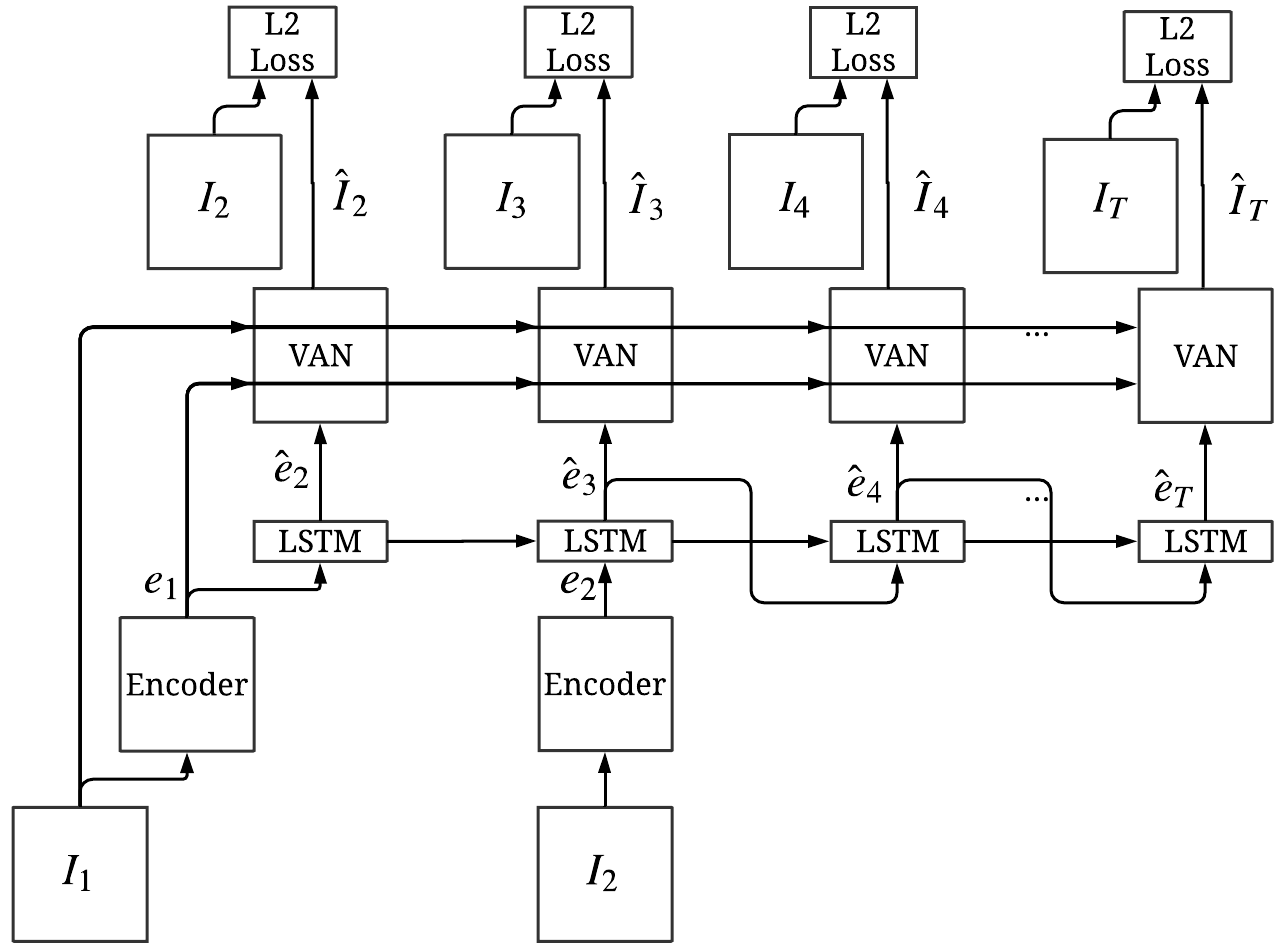}
\end{center}
\vspace{-0.15in}
\caption{The \textsc{E2E} method. The first few frames are encoded and fed into the predictor as context. The predictor predicts the subsequent encodings, which the VAN uses to produce the pixel-level predictions. The average of the losses is minimized. This is the configuration of every method at inference time, even if the predictor and VAN are trained separately.}
\label{fig:e2e_diagram}
\vspace{-0.1in}
\end{figure}

One simple baseline method is to simply connect the VAN and the predictor LSTM together, and train them end-to-end (E2E).
Our full network is optimized to minimize the L2 loss between the predicted image and the ground truth by: $$\min( \, \sum_{t=1}^{T}L_{2}(\hat{I}_{t}, I_{t}) \, ).$$
Figure~\ref{fig:e2e_diagram} illustrates a diagram of this training scheme.
Although a straightforward objective function is optimized, minimizing the L2 loss directly on the image outputs from previous observations tends to produce blurry predictions.
This phenomenon has also been observed in several previous works~\cite{Mathieu16, 2017arXiv170405831V, VillegasICLR17}. 

\begin{figure*}[h!]
    \centering
    \includegraphics[width=0.9\linewidth]{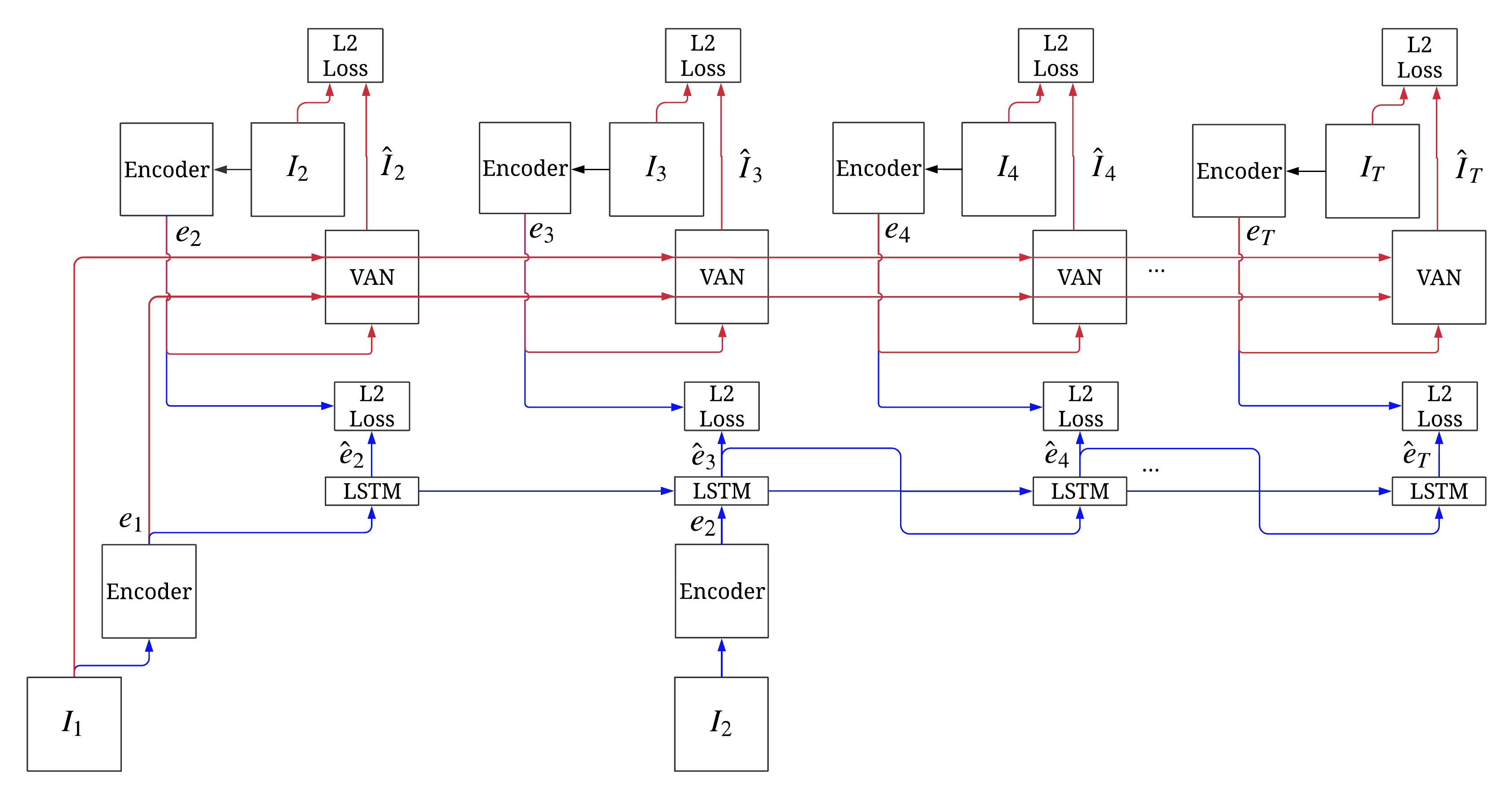}
    \vspace{-.1in}
    \caption{Blue lines represent the segment of the \textsc{EPVA} method in which the encoder and predictor are trained together. The encoder is trained to produce an encoding that is easy to predict, and the predictor is trained to predict that encoding into the future. Red lines represent the segment of the \textsc{EPVA} method in which the encoder and the VAN are trained together. The encoder is trained to produce an encoding that is informative to the VAN, while the VAN is trained to output the image given the encoding. The average of the losses in the diagram is minimized. This part of the method is similar to an autoencoder. Our method code is available at \url{https://bit.ly/2HqiHqx}}
    \label{fig:Full_EPVA_diagram}
    \vspace{-.2in}
\end{figure*}

\cutsubsubsectionup
\subsubsection{Encoder Predictor with Analogy Making}
\label{sec:EPVA}
\cutsubsubsectiondown

An alternative way to train our network is to constrain the features predicted by LSTM to be close to the outputs of the feature encoder (i.e. $\hat{e}_{t} \approx  e_{t}$).
Simultaneously, the feature encoder outputs can be trained to be useful for analogy making.
To accomplish this, we optimize the following objective function:
\begin{equation}
    \min ( \, \sum_{t=1}^{T}L_{2}(\hat{I}_t, I_t) + \alpha L_2(\hat{e}_t, e_t) \, ), \\
\end{equation}
where $\hat{I}_t = \mathit{VAN}(e_1, e_t, I_1)$, $e_t$ and $e_1$ are both outputs of the feature encoder computed from the image at time $t$ and the first image in the video, and $\alpha$ is a balancing hyper parameter that controls the importance between predicting $\hat{e}_t$ that is close to $e_t$ and learning an encoding $e_t$ that is good enough for image analogy.
$\alpha$ is used to prevent the predictor and encoder from both outputting the zero feature vector.

Figure~\ref{fig:Full_EPVA_diagram} illustrates the flow of information by which the encoder and predictor are trained together with blue arrows, and the flow of information by which the VAN and encoder are trained together with red arrows.
Separate gradient descent procedures (or optimizers, in TensorFlow parlance) could be used to minimize $L_{2}(\hat{I}_{t}, I_t)$ and $L_2(\hat{e}_t, e_t)$, but we found that minimizing the sum is more accurate in our experiments.
With this method, the predictor will generate the encoder outputs in future time steps, and the VAN will use the encoder output to produce the frame.
The advantage of this training scheme is that the VAN learns to sharply predict the pixels since it is trained given the encoding from the ground truth frame. The predictor learns to approximate the ground truth high-level features from the encoder.
Therefore, at inference time the VAN knows how to decode the high-level structure features resulting in better predictions.
Note that the encoder outputs $e_t$ are given to VAN as input during training; however, the predictor outputs $\hat{e}_t$ are given during testing. We refer to this method as EPVA.

The \textsc{EPVA} method works most accurately when experimented with $\alpha$ starting small, around 1e-7, and gradually increased to around $0.1$ during training. As a result, the encoder will first be optimized to produce an informative encoding, then gradually optimized to make that encoding easy to predict by the predictor.
\begin{table*}[t!]
\caption{Crowd-sourced human preference evaluation on the moving shapes dataset.}
\label{tab:shape}
\begin{center}
\begin{tabular}{lccc}
\multicolumn{1}{c}{\bf Method}  &\multicolumn{1}{c}{\bf Shape has correct color} &\multicolumn{1}{c}{\bf Shape has wrong color} &\multicolumn{1}{c}{\bf Shape disappeared}
\\ \hline
EPVA            &96.9\%    &3.1\%     &0\%      \\
CDNA Baseline   &24.6\%    &5.7\%     &69.7\%    \\
\end{tabular}
\end{center}
\vspace{-.1in}
\end{table*}

\begin{figure*}[!t]
    \vspace{-6pt}
    \centering
	\begin{subfigure}{0.04\linewidth}
        \raggedleft
        \rotatebox{90}{
        \hspace{-.4cm} \parbox{2cm}{\centering G.T.} \hspace{-.8cm} \parbox{2cm}{\centering \citet{2016arXiv160507157F}} \hspace{-.8cm} \parbox{2cm}{\centering EPVA}
        }
    \end{subfigure}
    \begin{subfigure}{0.08\linewidth}
        \caption*{t=1}
        \vspace{-7pt}
  		\includegraphics[width=1\linewidth]{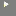}
  		\includegraphics[width=1\linewidth]{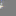}
  		\includegraphics[width=1\linewidth]{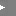}
	\end{subfigure} 
    \begin{subfigure}{0.08\linewidth}
        \caption*{t=2}
        \vspace{-7pt}
	    \includegraphics[width=1\linewidth]{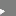}
	    \includegraphics[width=1\linewidth]{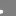}
  		\includegraphics[width=1\linewidth]{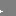}
	\end{subfigure} 
    \begin{subfigure}{0.08\linewidth}
        \caption*{t=3}
        \vspace{-7pt}
	    \includegraphics[width=1\linewidth]{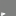}
	    \includegraphics[width=1\linewidth]{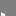}
  		\includegraphics[width=1\linewidth]{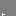}
	\end{subfigure} 
    \begin{subfigure}{0.08\linewidth}
        \caption*{t=256}
        \vspace{-7pt}
	    \includegraphics[width=1\linewidth]{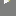} 
	    \includegraphics[width=1\linewidth]{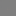}
  		\includegraphics[width=1\linewidth]{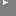}
	\end{subfigure}
	\begin{subfigure}{0.08\linewidth}
        \caption*{t=257}
        \vspace{-7pt}
	    \includegraphics[width=1\linewidth]{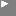} 
	    \includegraphics[width=1\linewidth]{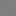}
  		\includegraphics[width=1\linewidth]{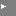}
	\end{subfigure}
	\begin{subfigure}{0.08\linewidth}
        \caption*{t=258}
        \vspace{-7pt}
	    \includegraphics[width=1\linewidth]{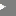}
	    \includegraphics[width=1\linewidth]{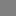}
  		\includegraphics[width=1\linewidth]{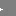}
	\end{subfigure}
	\begin{subfigure}{0.08\linewidth}
        \caption*{t=1020}
        \vspace{-7pt}
	    \includegraphics[width=1\linewidth]{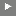}
	    \includegraphics[width=1\linewidth]{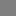}
  		\includegraphics[width=1\linewidth]{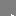}
	\end{subfigure}
	\begin{subfigure}{0.08\linewidth}
        \caption*{t=1021}
        \vspace{-7pt}
	    \includegraphics[width=1\linewidth]{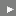}
	    \includegraphics[width=1\linewidth]{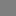}
  		\includegraphics[width=1\linewidth]{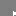}
	\end{subfigure}
	\begin{subfigure}{0.08\linewidth}
        \caption*{t=1022}
        \vspace{-7pt}
	    \includegraphics[width=1\linewidth]{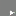}
	    \includegraphics[width=1\linewidth]{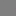}
  		\includegraphics[width=1\linewidth]{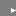}
	\end{subfigure}
    \vspace{-.1cm}
    \caption{A visual comparison of the \textsc{EPVA} method and CDNA from \citet{2016arXiv160507157F} as the baseline. This is a representative example of the quality of predictions from both methods. For videos please visit~\url{https://bit.ly/2kS8r16}.}
\label{fig:shapes_results}
\vspace{-.1in}
\end{figure*}

\cutsubsubsectionup
\subsubsection{EPVA with adversarial loss in predictor}
\cutsubsubsectiondown

A disadvantage of the EPVA training scheme alone is that the predictor is trained to minimize the L2 loss with respect to the encoder outputs.
The L2 loss is notoriously known for the ``blurriness effect," and it causes our predictor LSTM to output blurry predictions in encoding space.

One solution to this problem is to use an adversarial loss \cite{2014arXiv1406.2661G} between the predictor and encoder.
We use an LSTM discriminator network, which takes a sequence of encodings and produces a score that indicates whether the encodings came from the predictor or the encoder network.
We train the discriminator to minimize the improved Wasserstein loss \cite{ImprovedWGAN}.
\begin{equation}
\min ( \, \sum_{t=1}^{T}(D(\hat{e})- D(e) + \lambda (\left \| \nabla_{\hat{e}}D(\hat{e}) \right \|_2 -1)^2]) \, ).
\end{equation}
Here, $e$ and $\hat{e}$ are the sequence of inferred and predicted encodings respectively.
We train both the encoder and the predictor, 
so we use a loss which takes both the encoder and predictor outputs into account.
Therefore, we use the negative of the discriminator loss to optimize the generator.
\begin{equation}
\min ( \, -\sum_{t=1}^{T}(D(\hat{e}) - D(e)) \, )
\end{equation}
We also still optimize the l2 loss between the predictor and encoder, weighted by a scale factor. This ensures the predictions will be accurate given the context frame. We also feed a Gaussian noise variable into the predictor in order to generate different results given the same input sequence. We found that the noise helps generate more complex predictions in practice.

In addition to passing the predictor or encoder output to the discriminator, we also pass the output of the VAN encoder,  given the predictor or encoder output. This trains the predictor and encoder to encourage the VAN to produce similar quality images. This is achieved by substituting $[f_{enc}(e), e]$ for $e$ and $[f_{enc}(\hat{e}), \hat{e}]$ for $\hat{e}$ in the equations above, where $f_{enc}$ is the VAN encoder. 
The encoder and VAN are trained together in the same way as previously discussed.

\cutsectionup
\section{Experiments}
\label{sec:experiments}
\cutsectiondown

We evaluated our methods on two datasets: the Human 3.6M dataset \citep{h36m_pami,IonescuSminchisescu11}, and a toy dataset based on videos of bouncing shapes. 
More sample videos and code to reproduce our results are available at our project website \url{https://bit.ly/2kS8r16}.

\cutsubsectionup
\subsection{Long-term Prediction on a Toy Dataset}
\cutsubsectiondown

We train our method on a toy task with known factors of variation. We used a dataset with a generated shape that bounces around the image and changes size deterministically. We trained our \textsc{EPVA} method and the CDNA method from \citet{2016arXiv160507157F} to predict 16 frames, given the first three frames as context. Both methods are evaluated on predicting approximately 1000 frames. We added noise to the LSTM states of the predictor network during training to help predict accurate motion further into the future. Results from a held out test set are described in the following.

After visually analyzing the results of both methods, we found that when the CDNA fails, the shape disappears entirely. In contrast, when the \textsc{EPVA} method fails, the shape changes color. See Figure~\ref{fig:shapes_results} for sample predictions. 
For quantitative evaluation, we used a script to measure whether a shape was present from frames 1012 to 1022 and if that shape has the appropriate color. Table \ref{tab:shape} shows the results averaged over 1000 runs. The CDNA method predicts a shape with the correct color about 25\% of the time, and the EPVA method predicts a shape with the correct color about 97\% of the time. The EPVA method sometimes fails by predicting the shape in the same location from frame to frame, but this is rare as the reader can confirm by examining the randomly sampled predictions on our project website.
It is unrealistic to expect the methods to predict the location of the shape accurately in frame 1000 since small errors propagate in each prediction step.

\begin{table*}[t!]
\caption{Crowd-sourced human preference evaluation on the Human3.6M dataset.}
\label{tab:human_eval}
\begin{center}
\begin{tabular}{lccc}
\multicolumn{1}{c}{\bf Comparison}  &\multicolumn{1}{c}{\bf Ours is better}
&\multicolumn{1}{c}{\bf Same}
&\multicolumn{1}{c}{\bf Baseline is better}
\\ \hline
EPVA 1-127 vs \citet{2016arXiv160507157F} 1-127                                     &46.4\%         &40.7\%      &12.9\%\\
\textsc{EPVA Adv.} 1-127 vs \citet{2016arXiv160507157F} 1-127                &73.9\%         &13.2\%     &12.9\% \\ 
\textsc{EPVA Adv.} 63-127 vs \citet{2016arXiv160507157F} 1-63                &67.2\%     &17.5\%     &15.3\% \\
\textsc{EPVA Adv.} 5-127 vs \citet{Denton} 5-127 &58.2\%     &24.0\%      &17.8\%\\
\end{tabular}
\end{center}
\vspace{-.2in}
\end{table*}

\cutsubsectionup
\subsection{Long-term Prediction on Human3.6M}
\label{sec:humans_results}
\cutsubsectiondown

In these experiments, we use subjects 1, 5, 6, 7, and 8 for training, and subject 9 for validation. Subject 11 results are reported in this paper for testing. We use 64 by 64 images, and subsample the dataset to 6.25 frames per second. We train the methods to predict 32 frames and the results in this paper show predictions over 126 frames. Each method is given the first five frames as context. In these images, the model predicts about 20 seconds into the future starting with $0.8$ seconds of context. We use an encoding dimension of 64 for variations of our method on this dataset. 
The encoder in the \textsc{EPVA} method is initialized with the VGG network~\cite{Simonyan14c} pretrained on Imagenet \citep{imagenet_cvpr09}. 
To speed up the convergence of the \textsc{EPVA Adversarial} method, we start training from a pretrained \textsc{EPVA} model.

We compare our method to the CDNA method in \citet{2016arXiv160507157F} and the SVG-LP method in~\citet{Denton}. We trained each method with the same number of frames and context frames as ours. For~\citet{Denton}, we performed grid search on the $\beta$ and learning rate to find the best configuration for this experiment, as well as, used a network as large as we could fit in the GPU.
For~\citet{2016arXiv160507157F}, we performed grid search on the learning rate. The method in~\citet{Denton} can predict multiple futures, so we generate 5 futures for each context sequence, and compare against the one that most closely matches the ground truth in terms of SSIM. 
We find that this produces slightly better results than taking random predictions.
Note that this protocol provides an unfair advantage to their method.

\begin{figure*}[h!]
    \vspace{-6pt}
    \centering
	\begin{subfigure}{0.04\linewidth}
        \raggedleft
        \rotatebox{90}{
        \hspace{-.4cm} \parbox{2cm}{\centering \footnotesize Ground truth} \hspace{-.3cm} \parbox{2cm}{\centering \footnotesize \citet{2016arXiv160507157F}}
        \parbox{2cm}{\centering \footnotesize \citet{Denton}}
        \hspace{-.3cm} \parbox{2cm}{\centering \footnotesize Ours\\ Frames}
        \hspace{-.3cm} \parbox{2cm}{\centering \footnotesize Ours\\ Masks}
        }
    \end{subfigure}
    \begin{subfigure}{0.11\linewidth}
        \caption*{t=1}
        \vspace{-7pt}
        \includegraphics[width=1\linewidth]{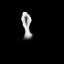}
  		\includegraphics[width=1\linewidth]{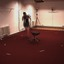}
  		\includegraphics[width=1\linewidth]{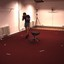}
  		\includegraphics[width=1\linewidth]{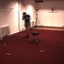}
  		\includegraphics[width=1\linewidth]{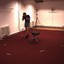}
	\end{subfigure} 
    \begin{subfigure}{0.11\linewidth}
        \caption*{t=36}
        \vspace{-7pt}
        \includegraphics[width=1\linewidth]{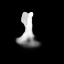} 
	    \includegraphics[width=1\linewidth]{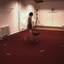} 
	    \includegraphics[width=1\linewidth]{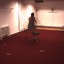}
  		\includegraphics[width=1\linewidth]{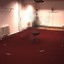}
  		\includegraphics[width=1\linewidth]{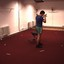}
	\end{subfigure} 
    \begin{subfigure}{0.11\linewidth}
        \caption*{t=54}
        \vspace{-7pt}
        \includegraphics[width=1\linewidth]{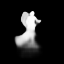}
	    \includegraphics[width=1\linewidth]{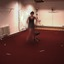} 
	    \includegraphics[width=1\linewidth]{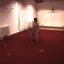}
  		\includegraphics[width=1\linewidth]{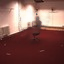}
  		\includegraphics[width=1\linewidth]{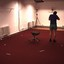}
	\end{subfigure} 
    \begin{subfigure}{0.11\linewidth}
        \caption*{t=72}
        \vspace{-7pt}
        \includegraphics[width=1\linewidth]{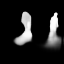}
	    \includegraphics[width=1\linewidth]{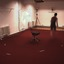} 
	    \includegraphics[width=1\linewidth]{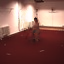}
  		\includegraphics[width=1\linewidth]{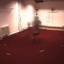}
  		\includegraphics[width=1\linewidth]{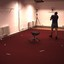}
	\end{subfigure}
	\begin{subfigure}{0.11\linewidth}
        \caption*{t=90}
        \vspace{-7pt}
        \includegraphics[width=1\linewidth]{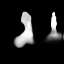}
	    \includegraphics[width=1\linewidth]{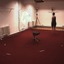} 
	    \includegraphics[width=1\linewidth]{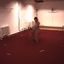}
  		\includegraphics[width=1\linewidth]{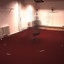}
  		\includegraphics[width=1\linewidth]{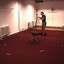}
	\end{subfigure}
	\begin{subfigure}{0.11\linewidth}
        \caption*{t=108}
        \vspace{-7pt}
        \includegraphics[width=1\linewidth]{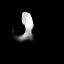}
	    \includegraphics[width=1\linewidth]{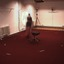} 
	    \includegraphics[width=1\linewidth]{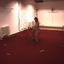}
  		\includegraphics[width=1\linewidth]{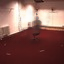}
  		\includegraphics[width=1\linewidth]{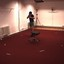}
	\end{subfigure}
	\begin{subfigure}{0.11\linewidth}
        \caption*{t=126}
        \vspace{-7pt}
        \includegraphics[width=1\linewidth]{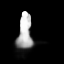}
	    \includegraphics[width=1\linewidth]{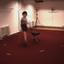} 
	    \includegraphics[width=1\linewidth]{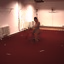}
  		\includegraphics[width=1\linewidth]{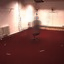}
  		\includegraphics[width=1\linewidth]{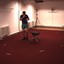}
	\end{subfigure}
    \vspace{-.0cm}\\
    \centering

	\begin{subfigure}{0.04\linewidth}
        \raggedleft
        \rotatebox{90}{
        \hspace{-0cm} \parbox{2cm}{\centering \footnotesize Ground truth}  \hspace{-.3cm} \parbox{2cm}{\centering \footnotesize \citet{2016arXiv160507157F}}
        \parbox{2cm}{\centering \footnotesize \citet{Denton}}
        \hspace{-.1cm}\parbox{2cm}{\footnotesize \centering Ours\\ Frames}
        \hspace{-.3cm} \parbox{2cm}{\footnotesize \centering Ours\\ Masks}
        }
    \end{subfigure}
    \begin{subfigure}{0.11\linewidth}
        \includegraphics[width=1\linewidth]{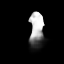}
  		\includegraphics[width=1\linewidth]{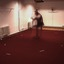}
  		\includegraphics[width=1\linewidth]{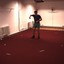}
  		\includegraphics[width=1\linewidth]{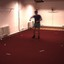}
  		\includegraphics[width=1\linewidth]{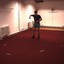}
	\end{subfigure} 
    \begin{subfigure}{0.11\linewidth}
        \includegraphics[width=1\linewidth]{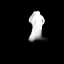}
	    \includegraphics[width=1\linewidth]{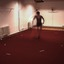} 
	    \includegraphics[width=1\linewidth]{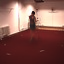}
  		\includegraphics[width=1\linewidth]{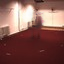}
  		\includegraphics[width=1\linewidth]{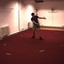}
	\end{subfigure} 
    \begin{subfigure}{0.11\linewidth}
        \includegraphics[width=1\linewidth]{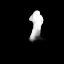}
	    \includegraphics[width=1\linewidth]{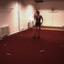} 
	    \includegraphics[width=1\linewidth]{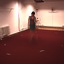}
  		\includegraphics[width=1\linewidth]{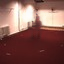}
  		\includegraphics[width=1\linewidth]{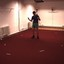}
	\end{subfigure} 
    \begin{subfigure}{0.11\linewidth}
        \includegraphics[width=1\linewidth]{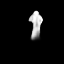}
	    \includegraphics[width=1\linewidth]{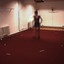} 
	    \includegraphics[width=1\linewidth]{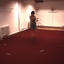}
  		\includegraphics[width=1\linewidth]{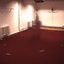}
  		\includegraphics[width=1\linewidth]{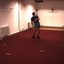}
	\end{subfigure}
	\begin{subfigure}{0.11\linewidth}
        \includegraphics[width=1\linewidth]{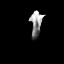}
	    \includegraphics[width=1\linewidth]{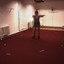} 
	    \includegraphics[width=1\linewidth]{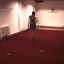}
  		\includegraphics[width=1\linewidth]{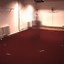}
  		\includegraphics[width=1\linewidth]{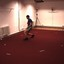}
	\end{subfigure}
	\begin{subfigure}{0.11\linewidth}
        \includegraphics[width=1\linewidth]{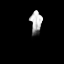}
	    \includegraphics[width=1\linewidth]{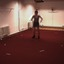} 
	    \includegraphics[width=1\linewidth]{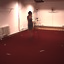}
  		\includegraphics[width=1\linewidth]{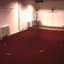}
  		\includegraphics[width=1\linewidth]{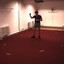}
	\end{subfigure}
	\begin{subfigure}{0.11\linewidth}
        \includegraphics[width=1\linewidth]{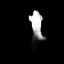}
	    \includegraphics[width=1\linewidth]{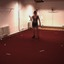} 
	    \includegraphics[width=1\linewidth]{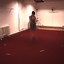}
  		\includegraphics[width=1\linewidth]{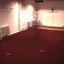}
  		\includegraphics[width=1\linewidth]{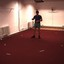}
	\end{subfigure}
    \vspace{-.1cm}\\
    \centering

    \caption{Comparison of the generated videos from \textsc{EPVA} with the \textsc{Adversarial loss} (ours), CDNA \cite{2016arXiv160507157F}, and SVG-LP \cite{Denton}.
    We let each method predict 127 frames and show the time steps indicated on top of the figure. The person completely disappears in all the predictions generated using \citet{2016arXiv160507157F}. For the SVG-LP method \cite{Denton}, the person either stops moving or almost vanishes into the background. The \textsc{EPVA} with \textsc{Adversarial loss} method produces sharp predictions in comparison to the baselines. Additionally, we show the discovered foreground motion segmentation mask that allows our network to delete the human in the input frame (static mask in the top example) and generate the human in the future frames (moving mask in the top example). Please refer to our project website for video results: \url{https://bit.ly/2kS8r16}.}
\label{fig:humans_results}
\end{figure*}

Figure~\ref{fig:humans_results} shows comparison to the baselines, 
and different variations of our method are compared in Figure~\ref{fig:ablative}.
In Figure~\ref{fig:humans_results}, we also show the discovered foreground motion segmentation mask from our method. This mask clearly shows that the feature embeddings from our encoder and predictor encode the rough location and outline of the moving human.

From visually analyzing the results, we found that the E2E and CDNA methods usually blur out very quickly. The EPVA method produces accurate predictions further into the future, but the figure sometimes disappears. The human predictions from the \textsc{EPVA Adversarial} method disappear less often and usually reappear in a later time step.

The CDNA \cite{2016arXiv160507157F} and the E2E methods produce blurry images because they are trained to minimize L2 loss directly. In the \textsc{EPVA} method, the predictor and VAN are trained separately. This prevents the VAN from learning to produce blurry images when the predictor is not confident. The predictions will be sharp as long as the predictor network outputs a valid encoding. The \textsc{EPVA Adversarial} method makes the predictor network more likely to produce a valid encoding since the discriminator is trained to produce valid predictions. 
We also observe that there is more movement in the \textsc{EPVA Adversarial} method.

\cutsubsubsectionup
\subsubsection{Person Detector Evaluation}
\cutsubsubsectiondown

\begin{figure}[t] 
\begin{center}
\includegraphics[width=1.0\linewidth]{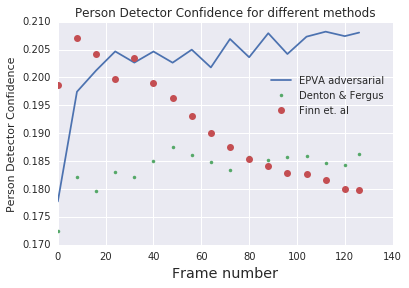}
\end{center}
\vspace{-.5cm}
\caption{Confidence of the person detector that a person is recognized in the predicted frame (``person score'').}
\vspace{-.2in}
\label{fig:person_score}
\end{figure}

We propose to compare the methods quantitatively by considering whether the generated videos contain a recognizable person. To do this in an automated fashion, we ran a MobileNet~\citep{mobilenet} object detection model pretrained on the MS-COCO~\citep{coco} dataset for each of the generated frames. We record the confidence of the detector that a person (one of the MS-COCO labels) is in the image. We call this the ``person score'' (with value ranges from 0 to 1, with a higher score corresponding to a higher confidence level). The human detector achieves approximately an accuracy of $0.4$ on the ground truth data. The results on each frame averaged over 1000 runs are shown in Figure \ref{fig:person_score}. 
The \textsc{EPVA Adversarial} method stays relatively constant over the different frames. For longer-term predictions, the evaluation shows that the \textsc{EPVA Adversarial} method is significantly better than the baselines.

\cutsubsubsectionup
\subsubsection{Human Evaluation}
\cutsubsubsectiondown

We also use a service similar to Mechanical Turk to collect comparisons of 1,000 generated videos from \citet{2016arXiv160507157F} and \citet{Denton} to different variations of our method. The task presents videos generated by the two methods side by side to human raters and asks them to confirm whether one of the videos is more realistic. The instructions tell raters to look for realistic motion, as well as a realistic person image. To evaluate the quality of the long-term predictions from the \textsc{EPVA Adversarial} method, we compare frames 64 to 127 of the \textsc{EPVA Adversarial} method to frames 1 to 63 of \citet{2016arXiv160507157F}. We evaluate frames 5-127 of \citet{Denton} against 5-127 of ours since their method isn't designed to produce good results for the context frames. 

The summary results are shown in Table \ref{tab:human_eval}. 
From these results, we conclude the following: the EPVA method generates significantly better long-term predictions than \citet{2016arXiv160507157F}. 
Further, the \textsc{EPVA Adversarial} method is a dramatic improvement over the EPVA method. The \textsc{EPVA Adversarial} method is capable of high-quality long-term predictions, as shown by frames 64 to 127 (seconds 10 to 20) of the \textsc{EPVA Adversarial} method being rated higher than frames 1-63 of \citet{2016arXiv160507157F}. The \textsc{EPVA Adversarial} is also significantly better than \citet{Denton} even after choosing the best out of 5 predictions after comparing with the ground truth in terms of SSIM.

\begin{figure*}[h]
    \vspace{-6pt}
    \centering
	\begin{subfigure}{0.04\linewidth}
        \raggedleft
        \rotatebox{90}{
        \hspace{-.4cm} \parbox{2cm}{\centering EPVA Adversarial} \hspace{-.4cm} \parbox{2cm}{\centering E2E and EPVA} \hspace{-.5cm} \parbox{2cm}{\centering EPVA} \hspace{-.3cm} \parbox{2cm}{\centering Without \\ VAN} \hspace{-.4cm}
        \parbox{2cm}{\centering E2E}
        }
    \end{subfigure}
    \begin{subfigure}{0.11\linewidth}
        \caption*{t=1}
        \vspace{-7pt}
        \includegraphics[width=1\linewidth]{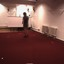}
        \includegraphics[width=1\linewidth]{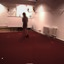}
  		\includegraphics[width=1\linewidth]{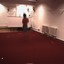}
  		\includegraphics[width=1\linewidth]{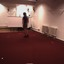}
  		\includegraphics[width=1\linewidth]{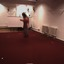}
	\end{subfigure} 
    \begin{subfigure}{0.11\linewidth}
        \caption*{t=36}
        \vspace{-7pt}
        \includegraphics[width=1\linewidth]{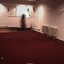}
        \includegraphics[width=1\linewidth]{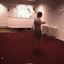}
	    \includegraphics[width=1\linewidth]{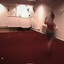} 
  		\includegraphics[width=1\linewidth]{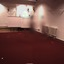}
  		\includegraphics[width=1\linewidth]{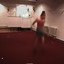}
	\end{subfigure} 
    \begin{subfigure}{0.11\linewidth}
        \caption*{t=54}
        \vspace{-7pt}
        \includegraphics[width=1\linewidth]{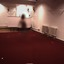}
        \includegraphics[width=1\linewidth]{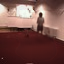}
	    \includegraphics[width=1\linewidth]{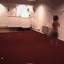} 
  		\includegraphics[width=1\linewidth]{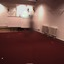}
  		\includegraphics[width=1\linewidth]{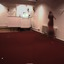}
	\end{subfigure} 
    \begin{subfigure}{0.11\linewidth}
        \caption*{t=72}
        \vspace{-7pt}
        \includegraphics[width=1\linewidth]{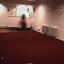}
        \includegraphics[width=1\linewidth]{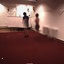}
	    \includegraphics[width=1\linewidth]{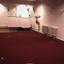} 
  		\includegraphics[width=1\linewidth]{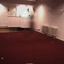}
  		\includegraphics[width=1\linewidth]{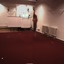}
	\end{subfigure}
	\begin{subfigure}{0.11\linewidth}
        \caption*{t=90}
        \vspace{-7pt}
        \includegraphics[width=1\linewidth]{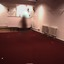}
        \includegraphics[width=1\linewidth]{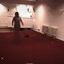}
	    \includegraphics[width=1\linewidth]{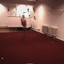} 
  		\includegraphics[width=1\linewidth]{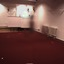}
  		\includegraphics[width=1\linewidth]{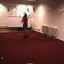}
	\end{subfigure}
	\begin{subfigure}{0.11\linewidth}
        \caption*{t=108}
        \vspace{-7pt}
        \includegraphics[width=1\linewidth]{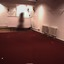}
        \includegraphics[width=1\linewidth]{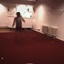}
	    \includegraphics[width=1\linewidth]{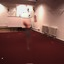} 
  		\includegraphics[width=1\linewidth]{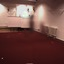}
  		\includegraphics[width=1\linewidth]{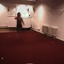}
	\end{subfigure}
	\begin{subfigure}{0.11\linewidth}
        \caption*{t=126}
        \vspace{-7pt}
        \includegraphics[width=1\linewidth]{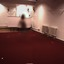}
        \includegraphics[width=1\linewidth]{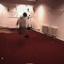}
	    \includegraphics[width=1\linewidth]{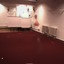} 
  		\includegraphics[width=1\linewidth]{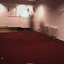}
  		\includegraphics[width=1\linewidth]{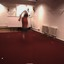}
	\end{subfigure}
    \vspace{-.15cm}\\
    \centering
    \hspace{-5pt}
    
    \caption{Ablative study illustration. We present comparisons between different variations of our architecture: E2E, loss without VAN, EPVA, combined E2E and EPVA loss, and our best model configuration (EPVA \textsc{Adversarial}). 
    See our project website for videos.
    }
\label{fig:ablative}
\vspace*{-0.15in}
\end{figure*}

\cutsubsubsectionup
\subsubsection{Pose regression from learned features}
\cutsubsubsectiondown

We perform experiments using the learned encoder features for human pose regression. We compare against a baseline based on features computed using the VGG network~\cite{Simonyan14c} trained for object recognition. The features are used as input to a 2-layer MLP, and trained to output human pose landmarks. The MLP trained with our features achieves an error of 0.0687 against an error of 0.0758 from the baseline features. This is a relative improvement of approximately $9\%$. This along with the generated masks shows the usefulness of our discovered features.

\cutsubsectionup
\subsection{Ablation Studies}
\cutsubsectiondown

We perform the following experiments to test different variations of the network and training.
We hypothesize that using a VAN improves the quality of the predictions.
To test this, we train a version of the network with the VAN replaced by a decoder network that only had access to the encoding and not the first observed frame. 

In this method, as well as the methods with the VAN, the decoder outputs a mask that controls whether to use its own output, or the pixels of the first frame. Thus, the decoder will have to set the mask values to not use the pixels from the first frame that correspond to the image of the person. Without the VAN, the network is often unable to set the mask values to completely remove the human from the first frame when predicting frames beyond 32. This is because the network is not always given access to the first frame, so it has to represent both foreground and background information in the prediction, which degrades over time.
Refer to Figure~\ref{fig:ablative} for comparison.

We also tried to use a hybrid objective that combines E2E and EPVA losses, but the videos generated from this method are more blurry than the videos from the EPVA method. These are called E2E and EPVA in Figure~\ref{fig:ablative}.
Finally, we also trained and evaluated the EPVA method with 10 frames of context instead of 5. We found that this didn't improve the long-term prediction results.

\cutsectionup
\section{Conclusion}
\label{sec:conclusion}
\cutsectiondown

We presented hierarchical long-term video prediction approaches that do not require ground truth high-level structure annotations. %
The proposed \textsc{EPVA} method has the limitation of the predictions occasionally disappearing, but it generates sharper images for a longer period of time compared to \citet{2016arXiv160507157F}, and the \textsc{E2E} method. 
By applying adversarial loss in the higher-level feature space, our \textsc{EPVA Adversarial} method generates more realistic predictions compared to all of the presented baselines including \citet{2016arXiv160507157F} and \citet{Denton}. 
This result suggests that it is beneficial to apply an adversarial loss in the higher-level feature space.
For future work, applying other techniques in feature space such as the variational method described in \citet{DBLP:journals/corr/abs-1710-11252} could enable our network to generate multiple future trajectories.

\vspace*{-0.3in}
\paragraph{Acknowledgments.}
We thank colleagues at Google Brain and anonymous reviewers for their constructive feedback and suggestions about this work. We also thank Emily Denton for providing her code available for comparison. 
R. Villegas was supported by Rackham Merit Fellowship.

\bibliography{submission}

\begin{thebibliography}{26}
\providecommand{\natexlab}[1]{#1}
\providecommand{\url}[1]{\texttt{#1}}
\expandafter\ifx\csname urlstyle\endcsname\relax
  \providecommand{\doi}[1]{doi: #1}\else
  \providecommand{\doi}{doi: \begingroup \urlstyle{rm}\Url}\fi

\bibitem[Babaeizadeh et~al.(2018)Babaeizadeh, Finn, Erhan, Campbell, and
  Levine]{DBLP:journals/corr/abs-1710-11252}
M.~Babaeizadeh, C.~Finn, D.~Erhan, R.~H. Campbell, and S.~Levine.
\newblock Stochastic variational video prediction.
\newblock In \emph{ICLR}, 2018.

\bibitem[Deng et~al.(2009)Deng, Dong, Socher, Li, Li, and
  Fei-Fei]{imagenet_cvpr09}
J.~Deng, W.~Dong, R.~Socher, L.-J. Li, K.~Li, and L.~Fei-Fei.
\newblock {ImageNet: A Large-Scale Hierarchical Image Database}.
\newblock In \emph{CVPR}, 2009.

\bibitem[Denton and Fergus(2018)]{Denton}
E.~Denton and R.~Fergus.
\newblock Stochastic video generation with a learned prior.
\newblock In \emph{ICML}, 2018.

\bibitem[Finn et~al.(2016)Finn, Goodfellow, and Levine]{2016arXiv160507157F}
C.~Finn, I.~Goodfellow, and S.~Levine.
\newblock Unsupervised learning for physical interaction through video
  prediction.
\newblock In \emph{NIPS}, 2016.

\bibitem[Goodfellow et~al.(2014)Goodfellow, Pouget-Abadie, Mirza, Xu,
  Warde-Farley, Ozair, Courville, and Bengio]{2014arXiv1406.2661G}
I.~Goodfellow, J.~Pouget-Abadie, M.~Mirza, B.~Xu, D.~Warde-Farley, S.~Ozair,
  A.~Courville, and Y.~Bengio.
\newblock Generative adversarial nets.
\newblock In \emph{NIPS}, 2014.

\bibitem[Goroshin et~al.(2015)Goroshin, Mathieu, and LeCun]{Goroshin15}
R.~Goroshin, M.~Mathieu, and Y.~LeCun.
\newblock Learning to linearize under uncertainty.
\newblock In \emph{NIPS}. 2015.

\bibitem[Gulrajani et~al.(2017)Gulrajani, Ahmed, Arjovsky, Dumoulin, and
  Courville]{ImprovedWGAN}
I.~Gulrajani, F.~Ahmed, M.~Arjovsky, V.~Dumoulin, and A.~C. Courville.
\newblock Improved training of wasserstein {GANs}.
\newblock In \emph{NIPS}, 2017.

\bibitem[Howard et~al.(2017)Howard, Zhu, Chen, Kalenichenko, Wang, Weyand,
  Andreetto, and Adam]{mobilenet}
A.~G. Howard, M.~Zhu, B.~Chen, D.~Kalenichenko, W.~Wang, T.~Weyand,
  M.~Andreetto, and H.~Adam.
\newblock {MobileNets}: Efficient convolutional neural networks for mobile
  vision applications.
\newblock \emph{arXiv preprint:1704.04861}, 2017.

\bibitem[Ionescu et~al.(2011)Ionescu, Li, and
  Sminchisescu]{IonescuSminchisescu11}
C.~Ionescu, F.~Li, and C.~Sminchisescu.
\newblock Latent structured models for human pose estimation.
\newblock In \emph{ICCV}, 2011.

\bibitem[Ionescu et~al.(2014)Ionescu, Papava, Olaru, and
  Sminchisescu]{h36m_pami}
C.~Ionescu, D.~Papava, V.~Olaru, and C.~Sminchisescu.
\newblock Human3.6m: Large scale datasets and predictive methods for 3d human
  sensing in natural environments.
\newblock \emph{IEEE Transactions on Pattern Analysis and Machine
  Intelligence}, 36\penalty0 (7):\penalty0 1325--1339, 2014.

\bibitem[Jayaraman and Grauman(2015)]{jayaraman2015learning}
D.~Jayaraman and K.~Grauman.
\newblock Learning image representations tied to ego-motion.
\newblock In \emph{ICCV}. 2015.

\bibitem[Jayaraman and Grauman(2016)]{jayaraman2016look}
D.~Jayaraman and K.~Grauman.
\newblock Look-ahead before you leap: end-to-end active recognition by
  forecasting the effect of motion.
\newblock In \emph{ECCV}, 2016.

\bibitem[Kalchbrenner et~al.(2017)Kalchbrenner, Oord, Simonyan, Danihelka,
  Vinyals, Graves, and Kavukcuoglu]{vpn}
N.~Kalchbrenner, A.~v.~d. Oord, K.~Simonyan, I.~Danihelka, O.~Vinyals,
  A.~Graves, and K.~Kavukcuoglu.
\newblock Video pixel networks.
\newblock In \emph{ICML}, 2017.

\bibitem[Lin et~al.(2014)Lin, Maire, Belongie, Bourdev, Girshick, Hays, Perona,
  Ramanan, Doll{\'{a}}r, and Zitnick]{coco}
T.~Lin, M.~Maire, S.~J. Belongie, L.~D. Bourdev, R.~B. Girshick, J.~Hays,
  P.~Perona, D.~Ramanan, P.~Doll{\'{a}}r, and C.~L. Zitnick.
\newblock Microsoft {COCO:} common objects in context.
\newblock \emph{ECCV}, 2014.

\bibitem[Lotter et~al.(2017)Lotter, Kreiman, and Cox]{Lotter17}
W.~Lotter, G.~Kreiman, and D.~Cox.
\newblock Deep predictive coding networks for video prediction and unsupervised
  learning.
\newblock In \emph{ICLR}. 2017.

\bibitem[Mathieu et~al.(2016)Mathieu, Couprie, and LeCun]{Mathieu16}
M.~Mathieu, C.~Couprie, and Y.~LeCun.
\newblock Deep multi-scale video prediction beyond mean square error.
\newblock In \emph{ICLR}. 2016.

\bibitem[Michalski et~al.(2014)Michalski, Memisevic, and
  Konda]{michalski_grammar_cells}
V.~Michalski, R.~Memisevic, and K.~Konda.
\newblock Modeling deep temporal dependencies with recurrent ``grammar cells".
\newblock In \emph{NIPS}, 2014.

\bibitem[Mittelman et~al.(2014)Mittelman, Kuipers, Savarese, and
  Lee]{icml2014c2_mittelman14}
R.~Mittelman, B.~Kuipers, S.~Savarese, and H.~Lee.
\newblock Structured recurrent temporal restricted {Boltzmann} machines.
\newblock In \emph{ICML}. 2014.

\bibitem[Oh et~al.(2015)Oh, Guo, Lee, Lewis, and Singh]{Oh15}
J.~Oh, X.~Guo, H.~Lee, R.~L. Lewis, and S.~Singh.
\newblock Action-conditional video prediction using deep networks in {Atari}
  games.
\newblock In \emph{NIPS}. 2015.

\bibitem[Ranzato et~al.(2014)Ranzato, Szlam, Bruna, Mathieu, Collobert, and
  Chopra]{Ranzato14}
M.~Ranzato, A.~Szlam, J.~Bruna, M.~Mathieu, R.~Collobert, and S.~Chopra.
\newblock Video (language) modeling: a baseline for generative models of
  natural videos.
\newblock \emph{arXiv preprint:1412.6604}, 2014.

\bibitem[Reed et~al.(2015)Reed, Zhang, Zhang, and Lee]{NIPS2015_5845}
S.~E. Reed, Y.~Zhang, Y.~Zhang, and H.~Lee.
\newblock Deep visual analogy-making.
\newblock In \emph{NIPS}. 2015.

\bibitem[Simonyan and Zisserman(2015)]{Simonyan14c}
K.~Simonyan and A.~Zisserman.
\newblock Very deep convolutional networks for large-scale image recognition.
\newblock In \emph{ICLR}, 2015.

\bibitem[Srivastava et~al.(2015)Srivastava, Mansimov, and
  Salakhudinov]{Srivastava15}
N.~Srivastava, E.~Mansimov, and R.~Salakhudinov.
\newblock Unsupervised learning of video representations using {LSTMs}.
\newblock In \emph{ICML}. 2015.

\bibitem[Sutskever et~al.(2009)Sutskever, Hinton, and Taylor]{NIPS2008_3567}
I.~Sutskever, G.~E. Hinton, and G.~W. Taylor.
\newblock The recurrent temporal restricted {Boltzmann} machine.
\newblock In \emph{NIPS}. 2009.

\bibitem[Villegas et~al.(2017{\natexlab{a}})Villegas, Yang, Hong, Lin, and
  Lee]{VillegasICLR17}
R.~Villegas, J.~Yang, S.~Hong, X.~Lin, and H.~Lee.
\newblock Decomposing motion and content for natural video sequence prediction.
\newblock In \emph{ICLR}. 2017{\natexlab{a}}.

\bibitem[Villegas et~al.(2017{\natexlab{b}})Villegas, Yang, Zou, Sohn, Lin, and
  Lee]{2017arXiv170405831V}
R.~Villegas, J.~Yang, Y.~Zou, S.~Sohn, X.~Lin, and H.~Lee.
\newblock Learning to generate long-term future via hierarchical prediction.
\newblock In \emph{ICML}, 2017{\natexlab{b}}.

\end{thebibliography}
\bibliographystyle{abbrvnat}


\end{document}